\documentclass[10pt,twocolumn,letterpaper]{article}

\usepackage{cvpr}              

\usepackage{graphicx}
\usepackage{amsmath}
\usepackage{amssymb}
\usepackage{booktabs}
\usepackage{algorithm}
\usepackage{algorithmic}
\usepackage{multirow}
\usepackage{bbding}
\usepackage{subcaption}
\usepackage{color}
\usepackage{caption} 
\usepackage{xcolor}

\usepackage[pagebackref,breaklinks,colorlinks]{hyperref}

\usepackage[capitalize]{cleveref}
\crefname{section}{Sec.}{Secs.}
\Crefname{section}{Section}{Sections}
\Crefname{table}{Table}{Tables}
\crefname{table}{Tab.}{Tabs.}

\def\cvprPaperID{11202} 
\def\confName{ICCV}
\def\confYear{2023}

\begin{document}

\title{Searching for the Fakes: Efficient Neural Architecture Search \\ 
	for General Face Forgery Detection}

\author{Xiao Jin, Xin-Yue Mu, Jing Xu\\
College of Artificial Intelligence, Nankai University\\
38 Tongyan Road, Jinnan District, Tianjin, P.R.China 300350\\
{\tt\small \{jinxiao,xujing\}@nankai.edu.cn}
}
\maketitle

\begin{abstract}
   As the saying goes, ``\textit{seeing is believing}". However, with the development of digital face editing tools, we can no longer trust what we can see. Although face forgery detection has made promising progress, most current methods are designed manually by human experts, which is labor-consuming. In this paper, we develop an end-to-end framework based on neural architecture search (NAS) for deepfake detection, which can automatically design network architectures without human intervention. First, a forgery-oriented search space is created to choose appropriate operations for this task. Second, we propose a novel performance estimation metric, which guides the search process to select more general models. The cross-dataset search is also considered to develop more general architectures. Eventually, we connect the cells in a cascaded pyramid way for final forgery classification. Compared with state-of-the-art networks artificially designed, our method achieves competitive performance in both in-dataset and cross-dataset scenarios.
\end{abstract}

\section{Introduction}
\label{sec:intro}
The emergence of face forgery technology has caused serious concerns in the whole society \cite{Beridze2019NMI,Hine2022NMI}.
First, using photos of political figures, stars and other people may incur legal issues, such as reputations, portrait rights, intellectual properties and so on. 
Second, videos of important persons are highly valuable intelligence.
The abuse of counterfeit technology can lead to the spread of false information, even causes diplomatic crisis.

\textit{In this paper, we mainly focus on two challenges of face forgery detection. The first one is the limitation of the performance in cross-dataset settings.} Existing methods often suffer from performance degradation when the training set is different from the test set. The forgery tools and video contents can vary significantly between two datasets. Forgery tools have diverse principles and processes, leaving distinct clues in the video. 
In addition, video content also affects the practicability of some existing techniques.
Due to the difference in skin colors and facial features of persons, the performance of classifiers may decrease to a certain degree \cite{Hu2022AAAI,Huang2021AAAI}.

There are some prior arts for general face forgery detection. The previous approaches mainly aim to extract more general features \cite{Yu2022TOMM,Luo2021CVPR,Yu2022TIFS}. However, feature extraction is only a part of a neural network. How to design a general detection framework as a whole has attracted relatively much less attention. Additionally, till now, most networks for face forgery detection are manually designed, which are labor expensive. 

\textit{The second challenge is the limitation of manually designed networks.}
Artificially designed neural networks have achieved good performance, while it requires rich professional knowledge and a lot of repeated experiments. As the number of network layers increases, more hyperparameters need to be manually adjusted, which consumes labor, time, and computing resources. This situation limits the further development of this field to some extent. The emergence of neural architecture search (NAS) technology provides a new possibility to solve this problem.

Although numerous NAS algorithms have been proposed, the inherent difference between face forgery detection and other computer vision tasks makes it difficult to adopt other NAS methods directly. First, the manipulation traces are too fragile to be discovered by ordinary layers in the neural networks. Second, some NAS methods are prone to overfitting, while generalization capability is of importance for facial manipulation detection. Third, increasing the number of operations in the candidate space causes an inefficient search process. Liu et al. \cite{Liu2021automated} develop an automated deepfake detection framework using common convolutional layers as the search space. However, ordinary convolution is not specially designed to extract forgery traces, and there is still room for performance improvement if the search space is appropriately selected.

In summary, our contributions are as follows:
\begin{itemize} 
	\item  We propose a novel end-to-end NAS framework to automatically design a network architecture for face forgery detection. To better fit this task, we create the forgery-oriented search space. To the best of our knowledge, it is the first NAS-based method with specialized search space for deepfake forensics.
	\item In addition, we also introduce a novel performance estimation strategy to select more general architectures. Based on the cells obtained by NAS, we further develop a cell cascaded pyramid network (C2PN) for final face forgery detection, which incorporates multiscale features for performance improvements. 
	\item Extensive experiments and visualizations demonstrate the effectiveness of the proposed method on four benchmark datasets.
\end{itemize}

\section{Related Works}
\subsection{Face Forgery Detection} Some prior arts are proposed for face forgery detection \cite{Mirsky2021CSUR,Tolosana2020IF,Verdoliva2020JSTSP}. 
Nirkin et al. \cite{Nirkin2021TPAMI} introduce a method for detecting fake videos using a single frame. This method includes a face recognition network and a context recognition network for facial features.
Yang et al. \cite{Yang2021TIFS} utilize central difference convolution and atrous spatial pyramid pooling (ASPP) to propose a multi-scale texture difference model for face forgery detection. The central difference convolution combines pixel intensity and pixel gradient information to fully describe the texture clues. The ASPP preserves multi-scale information and prevents texture features from being corrupted.
Huang et al. \cite{Huang2021AAAI} develop an active defense framework to degrade the performance of malicious user-controlled face forgery models.
Zhou et al. \cite{Zhou2021CVPR} design an algorithm to solve the task of multi-person face forgery detection. Supervised only by video-level labels, the scheme explores multi-instance learning and focuses on tampered faces. Nevertheless, as the structure of neural networks becoming more and more complex, the hyperparameters that need to be adjusted increase geometrically. Designing networks manually for deepfake detection becomes labor-consuming due to repetitive experiments.

\subsection{Neural Architecture Search}
The NASNet \cite{Zoph2018CVPR} is the first algorithm to use modular search, which greatly reduces the computation time due to normal cells and reduction cells. 
The ENAS \cite{Pham2018ENAS} samples subgraphs from a large hypergraph, and the operations in the subgraph inherit the weights on the hypergraph. This approach facilitates the sharing of weights between different structures and greatly reduces the training time of neural network structures in performance prediction. 
The ProxylessNAS \cite{Cai2018ProxylessNas} searches based on MobileNetV2, changing the number of convolutional layers, convolutional kernel sizes and the number of filters in its internal MBConv unit. All MBConv units have no shared structure and therefore have more flexibility. The MnasNet \cite{Tan2019Mnasnet} formulates the search process as a multi-objective optimization problem, considering both model accuracy and inference latency. 
The DARTS \cite{Liu2018DARTS} defines the search space as a continuous space. Each searched cell is derivable and can be optimized by stochastic gradient descent. 
However, other NAS frameworks can not be adopted for face forgery detection directly because of the intrinsic discrepancy between deepfake detection and other computer vision tasks. How to analyze fake videos from a NAS-based perspective has received relatively much less attention. Some approaches based on NAS have been proposed for image inpainting detection \cite{Wu2022TCSVT}. However, they only use NAS for designing an extraction block. The forgery clues are still discovered by manually designed layers, which is not an end-to-end NAS framework. 

\section{The Proposed Method}
\begin{figure*}[t]
	\centering
	\includegraphics[width=.95\textwidth]{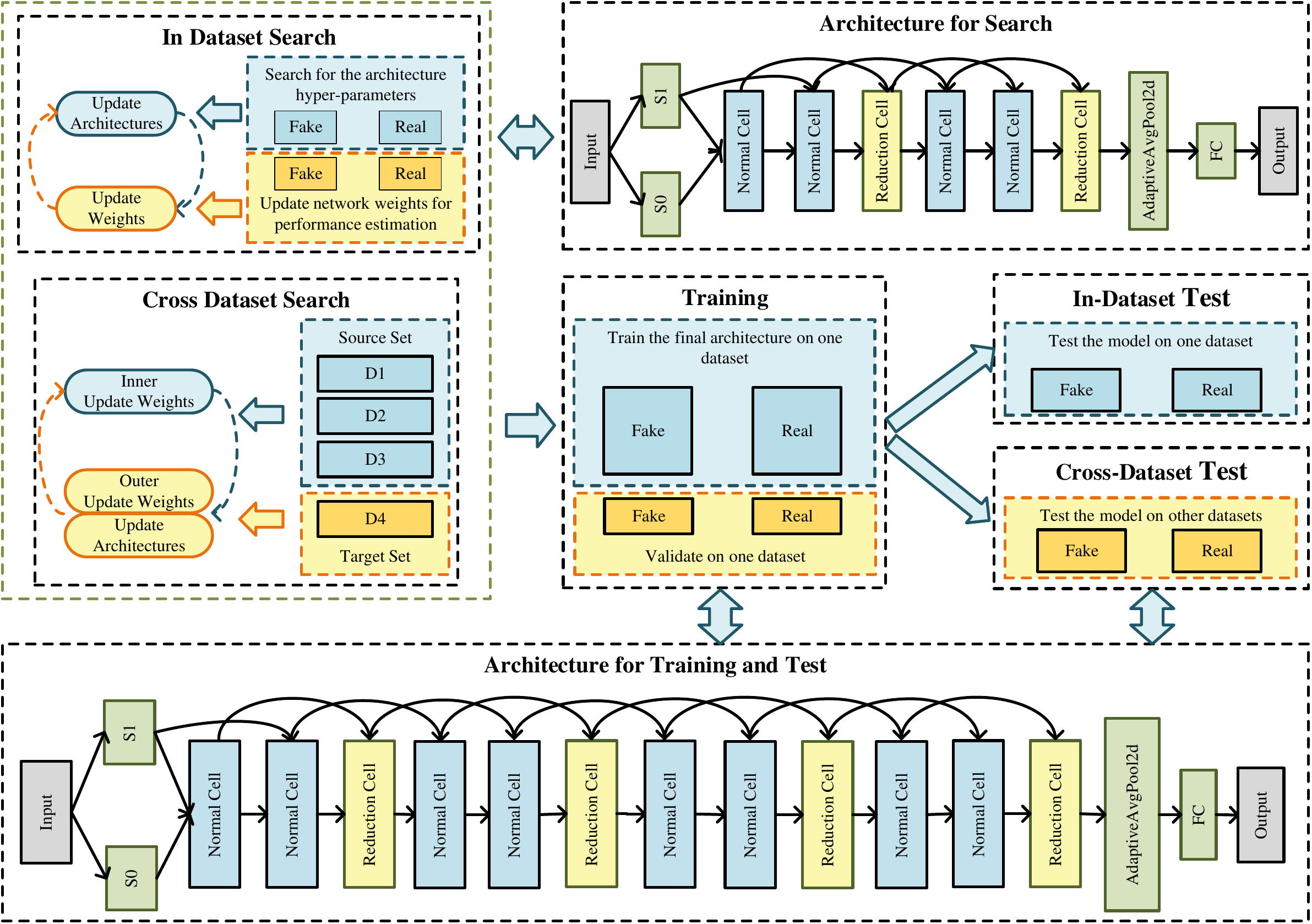}
	\caption{The framework of the proposed method. There are three main steps in our approach: search, training and test. In the search step, differentiable architecture search updates the architecture hyper-parameter and the network weights alternately. Due to the limitation of GPU memory, we adopt a lightweight network in this step. In training and test steps, we employ a deeper network to preserve valuable multiscale information for face forgery detection.}
	\label{fig:FRAMEWORK}
	\vspace{-.245in}
\end{figure*}

In this section, we first revisit the differentiable architecture search process. Then, we describe the search space, cell topology, performance estimation strategy, search strategy, and cross-dataset search, respectively. Eventually, we suggest the C2PN for final detection. The whole framework is illustrated in Fig. \ref{fig:FRAMEWORK}.

\subsection{Revisit Differentiable Architecture Search}
The process of neural architecture search mainly includes three parts: search space, search strategy and performance estimation. First, the search space defines all the candidate neural network operations or components. Then, the corresponding search strategy selects a certain neural network structure from the space. This intermediate derived network is evaluated by several performance indicators. Finally, the evaluation results are fed back to the search strategy until the optimal architectures are obtained.

Differentiable architecture search (DARTS) regards operation selection as weighted combination of operator sets, so that the search space is relaxed to be continuous \cite{Liu2018DARTS}. In this way, the hyperparameters can be calculated differentiably by gradient descent. 
The basic component in the architecture is a cell, which can be considered as a directed acyclic graph (DAG) with $N$ nodes. A topological node represents for a layer in the neural network. Usually, each cell is composed of two input nodes, one output node and the remaining intermediate nodes. For the $j$-th intermediate node, its output is a summation of all the forward results of preceding nodes,
\begin{equation}
\mathbf{x}_{j}=\sum_{i< j}f_{i, j}\left(\mathbf{x}_{i}\right), ~ 0 \leq i \le j \leq N-1,
\end{equation}
where $\mathbf{x}_{i}$ symbolizes the output feature map from the $i$-th node, $f_{i, j}(\cdot)$ is the operation from the $i$-th node to the $j$-th node.
In a cell, the directed edge between two nodes defines an operation, e.g., convolution, pooling, and activation function, selected from the search space $\mathcal{O}$. These operations can be formulated as,
\begin{equation}
f_{i, j}\left(\mathbf{x}_{i}\right)=\sum_{o \in \mathcal{O}} \frac{\exp \left\{\alpha_{i, j}^{o}\right\}}{\sum_{o^{\prime} \in \mathcal{O}} \exp \left\{\alpha_{i, j}^{o^{\prime}}\right\}} \cdot o\left(\mathbf{x}_{i}\right),
\end{equation}
where $o\left(\cdot\right)\in \mathcal{O}$ indicates candidate operations, $\alpha_{i, j}^{o}$ is a trainable variable that determines the network architecture.
Thus, the task of searching the network architecture is converted to optimizing all the $\alpha_{i, j}^{o}$ in the network.
After the search process, every node is connected with two preceding nodes that have maximum $\alpha_{i, j}^{o}$ value. The final output architecture is constructed by several cells. These cells can be divided into two types, i.e., the normal cell that preserves the spatial resolution, and reduction cell that decreases the resolution. 

\subsection{Search Space}
\begin{figure}[t]
	\centering
	\includegraphics[width=0.4\textwidth]{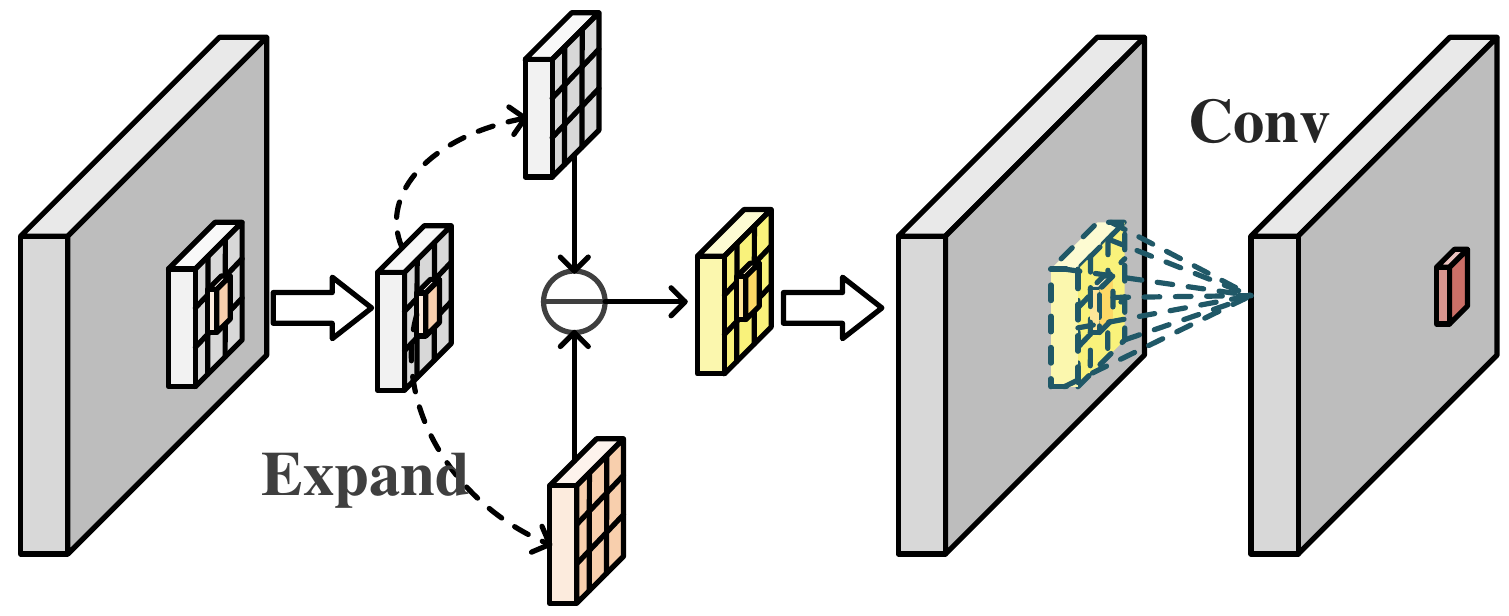}
	\caption{The structure of central difference convolution.}
	\label{fig:CDC}
	\vspace{-.2in}
\end{figure}

In this paper, the search space is mainly composed of Central Difference Convolution (CDC). 
Central difference operations have shown their effectiveness for face forgery detection. The features extracted with central difference can enhance the discriminability for forged images \cite{Yang2021TIFS}. Fig. \ref{fig:CDC} shows the structure of CDC. According to \cite{Yang2021TIFS,Yu2020CVPR}, the input is denoted as $z$, the output is represented by $y$, the process of CDC can be formulated as, 
\begin{equation}
	\begin{split}
	&y\left(p_0\right)=\\
	&\underbrace{\sum_{p_n \in \mathcal{R}} w\left(p_n\right) \cdot z\left(p_0+p_n\right)}_{\text {vanilla convolution }}+\theta \cdot \underbrace{\left(-z\left(p_0\right) \cdot \sum_{p_n \in \mathcal{R}} w\left(p_n\right)\right)}_{\text {central difference term }},
	\end{split}
\end{equation}
where $\mathcal{R}$ stands for local receptive field region, $w$ is the weight, $p_0$ is the current location on feature maps, $p_n$ is the other locations in $\mathcal{R}$, hyperparameter $\theta \in [0, 1]$ means the importance of gradient information.

\begin{figure*}[t]
	\centering
	\includegraphics[width=1\textwidth]{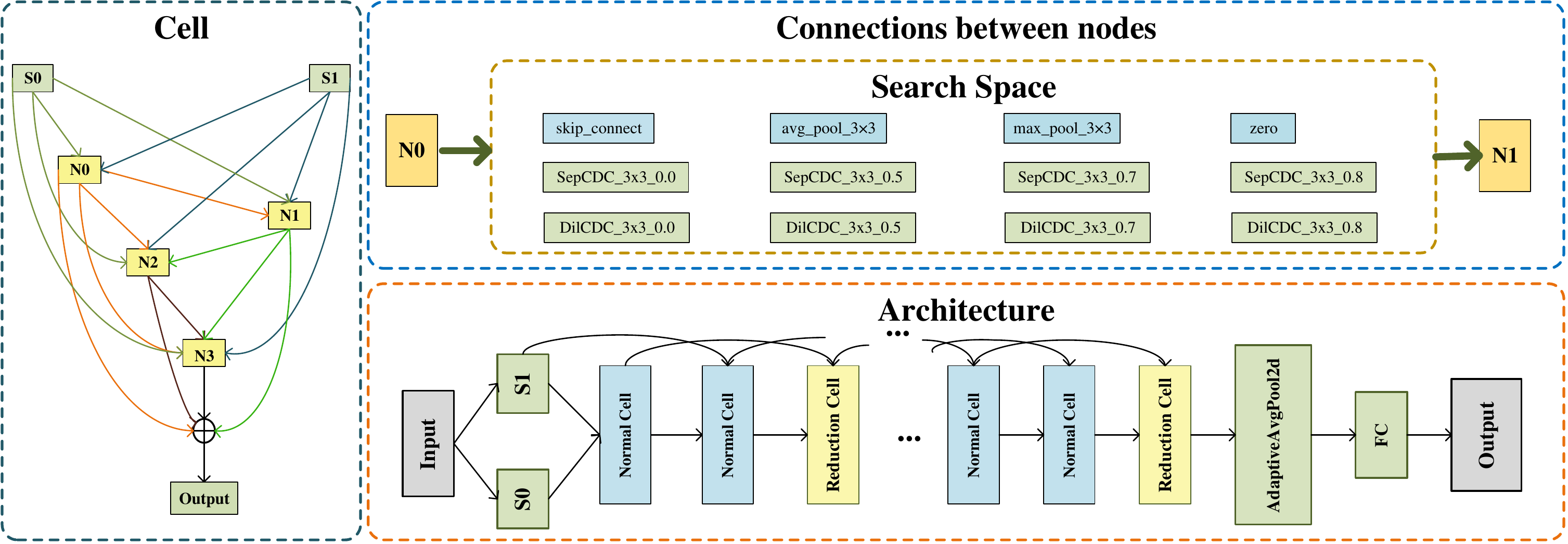}
	\caption{The illustration of the NAS. (a) The connections between nodes in a cell. Each edge indicates an operation obtained from the search space; (b) The topological structure of a cell; (c) The architecture of the whole detection networks.}
	\label{fig:NAS}
	\vspace{-.1in}
\end{figure*}

In this type of operation space, we involve eight central difference operations with various parameter settings as shown in Fig. \ref{fig:NAS}(a). ``Sep" denotes ``separable", while ``Dil" represents for ``dilated". For example, ``SepCDC\_3x3\_0.5" is defined as a separable central difference convolutional layer, where the kernel size is 3$\times$3, and $\theta$ is 0.5. 

When $\theta=0$, central difference operations are converted to vanilla CNN. The vanilla CNN operations aim to discover some visual forgery artifacts, e.g., uncommon manipulated region boundaries. Concretely, since the multiscale information plays an important role in deepfake detection \cite{Yang2021TIFS}, we emphasize the skip connection and dilated convolution in the search space, which are denoted as ``skip\_connect" and ``DilCDC\_3$\times$3", respectively.

In preliminary experiments, we also analyze the effectiveness of some noise extractors, e.g., Bayar layer \cite{Bayar2018TIFS} and SRM layer \cite{Fridrich2012TIFS}. We find central difference operations perform better.


\subsection{Cell Topology}
The topology structure of a cell is illustrated in Fig. \ref{fig:NAS}(b), where ``C" symbolizes concatenation.
In our method, each cell contains seven nodes, including two input nodes, e.g., S0 and S1, four intermediate nodes, e.g., N0, N1, N2, and N3, and one output node.
The edges indicate the operations searched from the operation space. 
At the beginning of the search, every node is linked with all predecessor nodes.
At the end of the search, every node only keeps connection with the two antecedent nodes with the highest scores. Finally, the search process produces two types of cells. The normal cells maintain the same resolution, while the reduction cells decrease the resolution of feature maps by half.

\subsection{Performance Estimation Strategy}
In previous gradient-based NAS methods, the significance of operations is ordered by architecture parameters $\alpha=\{\alpha^o_{i,j},0 \leq i \le j \leq N-1\}$, which indicates the relative importance of an operation compared with other candidate operations. However, the operation with the highest value of $\alpha$ may not always bring the best performance \cite{Zhou2021TNNLS}. Furthermore, DARTS tends to overfitting \cite{Zela2019ICLR}, while the generalization ability on unseen datasets is significant to face forgery detection. 

Therefore, we introduce additional conditions to select suitable operations,
\begin{equation}
E^{o}_{i, j}=\frac{\exp (1/\|\bar E^o_{i,j}-\hat E^o_{i,j}\|_{-\infty})}{\sum_{o^{\prime} \in \mathcal{O}} \exp (1/\|\bar E^o_{i,j}-\hat E^o_{i,j}\|_{-\infty})},
\end{equation} 
where $\bar E^o_{i,j}$ is the classification error in the search stage, $\hat E^o_{i,j}$ is the error in performance estimation process. Here, $\| \cdot\|_{-\infty}$ denotes the negative infinite norm, which returns the smallest absolute values of all elements. In this way, we encourage the error in performance estimation to close to the error in search, improving the generalization capability of the network.
The total estimation indicator is defined as,
\begin{equation}\label{eq:IND}
I^{o}_{i, j}=\frac{\exp \left\{\alpha_{i, j}^{o}\right\}}{\sum_{o^{\prime} \in \mathcal{O}} \exp \left\{\alpha_{i, j}^{o^{\prime}}\right\}}+\lambda E^{o}_{i, j},
\end{equation}
where $\lambda=0.15$ is the balance parameter.

\subsection{Search Strategy}
To improve the efficiency of the search process, the feature maps are sampled by the proportion of $1/n$ in the channel dimension \cite{Xu2021TPAMI}. This process can be defined as,
\begin{equation}
\begin{split}
f_{i, j}\left(\mathbf{x}_{i};M_{i, j}\right)=\sum_{o \in \mathcal{O}} I_{i, j}^{o} \cdot o\left(M_{i, j} \odot \mathbf{x}_{i}\right)\\
+\left(1-M_{i, j}\right) \odot \mathbf{x}_{i}\hfill,
\end{split}
\end{equation}
where $\odot$ represents the Hadamard product, $M_{(i,j)}\in\{0,1\}$ is channel mask, 1 denotes this feature channel is taken, 0 indicates this channel is discarded.

In the preliminary experiments, we find that the search process tends to select several types of operations, while neglecting other kinds of operations. To increase the diversity of the operations in the cell, we further take the edge parameter $\beta_{i,j}$ into consideration. The output of the $j$-th intermediate node is defined as,
\begin{equation}
\mathbf{x}_{j}=\sum_{i<j} \frac{\exp \left\{\beta_{i, j}\right\}}{\sum_{i^{\prime}<j} \exp \left\{\beta_{i^{\prime}, j}\right\}} \cdot f_{i, j}\left(\mathbf{x}_{i}\right).
\end{equation}

The importance of the connections between edges is determined by $I_{i, j}^{o}$ and $\beta_{i,j}$. We define $H^o_{i,j}$ to describe this process,
\begin{equation}\label{eq:HIND}
H^o_{i,j}=\frac{\exp \left\{\beta_{i, j}\right\}}{\sum_{i^{\prime}<j} \exp \left\{\beta_{i^{\prime}, j}\right\}} \cdot I_{i, j}^{o}.
\end{equation}
The operation with the highest $H^o_{i,j}$ is adopted for final neural network architecture,
\begin{equation}
{f^*_{i,j}} = \mathop {\arg \max }\limits_{o \in \mathcal{O}} H _{i,j}^o,
\end{equation}
where $f^*_{i,j}$ denotes the final operation from the $i$-th node to the $j$-th node.

Moreover, to further accelerate the search procedure, we update the operation state $\mathcal{S}\in\{\texttt{True},\texttt{False}\}$ between two nodes according to $H^o_{i,j}$. \texttt{True} denotes this operation is retained for the following search, while \texttt{False} indicates this operation is discarded for the search in the next step. After every twenty epochs, we remove the two operations with the lowest scores, and increase the rate of channel sampling. The whole procedure is illustrated in Algorithm \ref{alg:algorithm}.

\subsection{Cross-dataset Search}
A key challenge for deepfake detection is the performance degradation in cross-dataset scenarios. In this paper, we try to analyze this problem by automatically searching a general architecture. Inspired by \cite{Yu2021TPAMI}, we adopt a cross-dataset search strategy. For simplicity, assuming we have $K$ datasets $\mathcal{D}=\{\mathcal{D}_1,\mathcal{D}_2,\cdots,\mathcal{D}_K\}$, the randomly selected samples in $(K-1)$ datasets are defined as the source datasets $\mathcal{D}^s$, while the rest one is denoted as the target dataset $\mathcal{D}^t$. Then, the sample batch examples $\mathcal{I}^s_i(i=1,2,\cdots,K-1)$ in every dataset of $\mathcal{D}^s$, and $\mathcal{I}^q$ in $\mathcal{D}^t$ are employed for inner-updated network weights and architecture optimization stage, respectively. The process of cross-dataset search is described in Algorithm \ref{alg:cross-dataset-search}. Compared with \cite{Yu2021TPAMI}, our work utilizes more flexible search space to fully explore the potential of cross-dataset search strategy.  

\subsection{Cell Cascaded Pyramid Network}
After obtaining the cells, we connect them in series to detect face forgery videos. Since there are a large number of connections between nodes at the beginning of the search phase, it is computationally consuming. We use a relatively lightweight network, which contains two groups of cells, for architecture search. Each group is composed of two normal cells and a reduction cell. In total, six cells are used for the search stage, which is illustrated at the top of Fig. \ref{fig:FRAMEWORK}. 

Multiscale information is vital in deepfake forensics \cite{Yang2021TIFS}. We try to preserve this valuable multiscale information for the final detection. Thus, we use a deeper network to gradually decode the feature maps. As shown in the bottom of Fig. \ref{fig:FRAMEWORK}, the detection network contains four groups of cells. Each group has two normal cells followed by a reduction cell. We adopt the cells obtained from the above search stage without any training weights. The detection network is trained on benchmark datasets for in-dataset and cross-dataset tests.

\begin{algorithm}[t]
	\caption{The proposed method}
	\label{alg:algorithm}
	\textbf{Input}: dataset $\mathcal{D}$; operation set ${\mathcal{O}}$\\
	\textbf{Output}: cell topology
	\begin{algorithmic}[1] 
		\STATE Split the dataset evenly into two parts, $\mathcal{D}_{S}$ for search, and $\mathcal{D}_{Eval}$ for evaluation;
		\STATE Randomly sample a super-network and tune the network weights in the first 10 epochs;
		\WHILE{not converged}
		\STATE Update architecture hyper-parameters on $D_{S}$;
		\STATE Update network weights on $D_{Eval}$;
		\STATE Update $\bar E^o_{i,j}$ and $\hat E^o_{i,j}$;
		\STATE Calculate operation importance $H^o_{i,j}$ by Eq. (\ref{eq:HIND});
		\STATE Update each operation state $\mathcal{S}$.
		\IF {epoch$>$19 and epoch $\%$$20$ $==$ $0$}
		\STATE Prune operations according to their state $\mathcal{S}$ on each edge;
		\STATE Update channel sampling rate $n = 2n$;
		\ENDIF
		\ENDWHILE
		\STATE \textbf{return} cell topology
	\end{algorithmic}
\end{algorithm}

\begin{algorithm}[t]
	\caption{The cross-dataset search process}
	\label{alg:cross-dataset-search}
	\textbf{Input}: dataset $\mathcal{D}=\{\mathcal{D}_1,\mathcal{D}_2,\cdots,\mathcal{D}_K\}$\\
	\textbf{Output}: cell topology
	\begin{algorithmic}[1] 
		\STATE Initialize weights and architecture;
		\WHILE{not converged}
		\STATE Randomly select $(K-1)$ datasets in $\mathcal{D}$ as source datasets $\mathcal{D}^s$, and the rest one as target dataset $\mathcal{D}^t$;
		\STATE Sample batch examples $\mathcal{I}^s_i(i=1,2,\cdots,K-1)$ in every dataset of $\mathcal{D}^s$, the examples $\mathcal{I}^t$ in $\mathcal{D}^t$;
		\FOR{each $\mathcal{I}^s_i$}
		\STATE Update weights in the $i$-th dataset;
		\ENDFOR
		\STATE Update weights using all learners’ loss on the $\mathcal{I}^t$;
		\STATE Update architecture on the $\mathcal{I}^t$;
		\ENDWHILE
		\STATE \textbf{return} cell topology
	\end{algorithmic}
\end{algorithm}

\section{Experiment}
\subsection{Experimental Setup}
\textbf{Datasets.} We evaluate our method on four publicly available datasets. FaceForensics++ (FF++) \cite{FF++} consists of four forgery approaches, e.g., DeepFakes (DF), Face2Face (F2F), FaceSwap (FS), and NeuralTextures (NT). This dataset covers two compression rates, e.g., high-quality (C23) and low-quality (C40). 
Celeb-DF \cite{Celeb-DF} contains 590 real and 5,639 forged videos totally from YouTube website. These clips are generated by advanced synthesis techniques.
WildDeepfake \cite{Wilddeepfake} collects 7,314 real-world face sequences from the Internet. The duration of video clip changes dramatically.
DFDC-preview \cite{DFDC-pre} is composed of more than 5,000 videos with various manipulation methods. The faces in the video may be partially forged.

In-dataset search only uses FF++ for searching neural architectures. Cross-dataset search adopts four datasets alternately. For in-dataset scenes, after determining the network structure, the models are trained and tested on the above four datasets, respectively. For cross-dataset experiments, we train the networks on FF++ dataset and test on other three datasets to verify the generalization ability. 
In FF++ dataset, following the official splits, we use 720 videos for training, 140 videos for validation, and 140 videos for test. Each video takes 20 frames for training and 10 frames for validation and test. For the other three datasets, we divided them into the training set, validation set and test set according to the ratio of 8:1:1.

\textbf{Evaluation Metrics.} We apply two widely used metrics, accuracy score (Acc), and area under the receiver operating characteristic curve (AUC) for evaluation.

\textbf{Compared Methods.} The proposed method is compared with 15 state-of-the-art face forgery detection frameworks, including FWA \cite{FWA}, Meso4 \cite{Mesonet}, MesoInception4 \cite{Mesonet}, Xception \cite{FF++}, Multi-Task \cite{Multi-task}, Capsule \cite{Capsule}, EfficientNet-B4 \cite{EfficientNet}, Multi-Attention \cite{Multi-attention}, RNN \cite{RNN}, LTW \cite{LTW}, FT-TS \cite{FT-TS}, ADD\cite{Liu2021automated}, F3-Net \cite{F3-Net}, FInfer \cite{Hu2022AAAI}, and RECCE \cite{RECCE}. We directly report the quantitative results in the published papers if accessible.


\textbf{Implementation Details.} The facial areas are extracted by the \texttt{face\_recognition} toolbox.  
Due to GPU memory limitations, the inputs are resized to 32$\times$32. For in-dataset search, a half of data in FF++ is used for searching architecture hyper-parameters, while the other half is used for updating network weights. For cross-dataset search, we randomly selects 2,000 samples from each dataset to participate in the search in each epoch. The search process uses the Adam optimizer \cite{ADAM} for 65 epochs. 
The batch size is set to 96. The initial number of channels is 16, and the initial rate of channel sampling is 1/8. The running time of search stage is about 3 hours on an NVIDIA 3080Ti GPU due to channel sampling and pruning. After obtaining the determined architecture, we train the model on FF++ or the other three datasets for 150 epochs. The input pictures are resized to 64$\times$64. The batch size is set as 48. We adopt the SGD optimizer \cite{SGD} with an initial learning rate of 0.025, a momentum of 0.9, a weight decay of 3e-4.

\begin{figure*}[t]
	\centering
	\includegraphics[width=.6\textwidth]{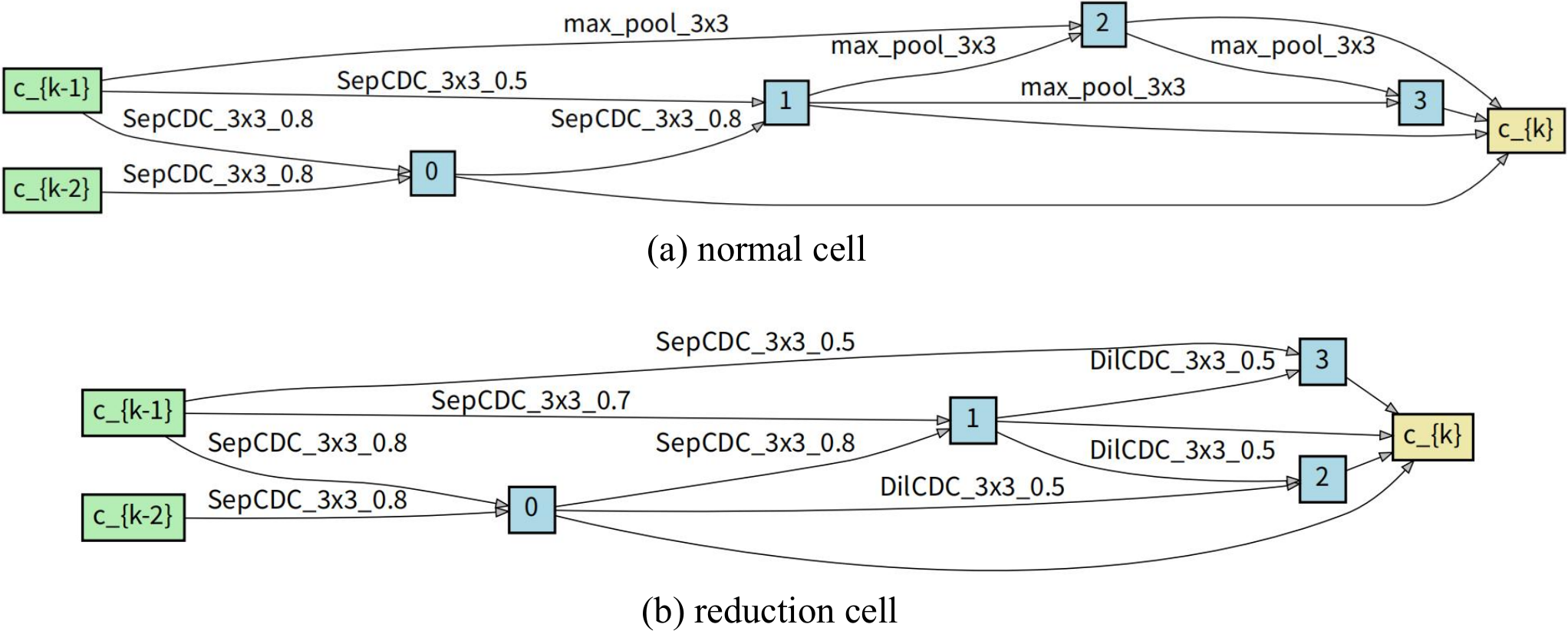}
	\caption{Detailed structure of the best cells discovered on FF++ by our in-dataset search. (a) normal cell. (b) reduction cell. 
	}
	\label{fig:cell}
	\vspace{-.05in}
\end{figure*}

\begin{figure*}[t]
	\centering
	\includegraphics[width=1\textwidth]{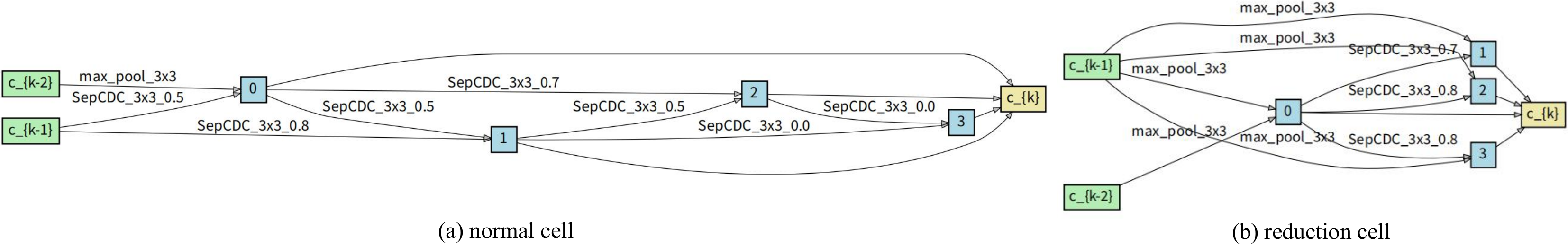}
	\caption{Detailed structure of the best cells discovered by our cross-dataset search. (a) normal cell. (b) reduction cell. 
	}
	\label{fig:cell-cross-domain}
	\vspace{-.05in}
\end{figure*}

\subsection{Comparison With State-of-the-art Methods}

\textbf{In-dataset Evaluation.} After the search process, the architecture is trained and tested on the above datasets, respectively. The results are shown in Table. \ref{tab:in-dataset} and the first column of Table. \ref{tab:cross-dataset1}. Except for FF++, the proposed method outperforms other baselines in this scenario. ``Ours$_{in}$" means that the search process only uses FF++, while ``Ours$_{cross}$" denotes cross-dataset search. 

\textbf{Cross-dataset Evaluation.} To validate the generalization capability, we train the network on FF++. Then, we test this model on other datasets. As listed in the second column of Table. \ref{tab:cross-dataset1} and Table. \ref{tab:cross-dataset2}, the proposed method achieves competitive performance with cross-dataset search. If the search stage only takes samples from FF++, our method still shows its effectiveness.

\begin{center}
	\setlength{\tabcolsep}{0.5mm}{
		\begin{table}[t]
			\small	
			\centering
			\caption{Comparisons of the in-dataset results on Celeb-DF, WildDeepfake, and DFDC-preview datasets.}
			\vspace{-.1in}	
			\label{tab:in-dataset}								
			\begin{tabular}{ccccccc} \hline										
				\multirow{2}{*}{Method} & \multicolumn{2}{c}{Celeb-DF} & \multicolumn{2}{c}{WildDeepfake} & \multicolumn{2}{c}{DFDC-pre} \\	
				& Acc & AUC & Acc & AUC & Acc & AUC \\ \hline									
				Meso4\cite{Mesonet} \textcolor{gray}{\scriptsize{[WIFS18]}} & 67.53 & 66.17 & 64.47 & 66.5 & 75.39 & 76.47 \\										
				RNN\cite{RNN} \textcolor{gray}{\scriptsize{[AVSS18]}} & 71.20 & 86.52 & 66.87 & 67.35 & 75.02 & 77.48 \\										
				FWA\cite{FWA} \textcolor{gray}{\scriptsize{[CVPRW19]}} & 64.73 & 60.16 & 55.46 & 57.92 & 73.25 & 72.97 \\										
				Xception\cite{FF++} \textcolor{gray}{\scriptsize{[ICCV19]}} & 90.34 & 89.75 & 75.26 & 80.89 & 79.32 & 81.58 \\										
				FT-TS\cite{FT-TS} \textcolor{gray}{\scriptsize{[TCSVT21]}} & 80.74 & 86.67 & 68.78 & 68.09 & 63.85 & 64.03 \\										
				FInfer\cite{Hu2022AAAI} \textcolor{gray}{\scriptsize{[AAAI22]}} & 90.47 & 93.30 & 75.88 & 81.38 & 80.39 & 82.88 \\
				\hline										
				Ours$_{in}$ & \textbf{94.92} & 99.59 & 78.25 & 85.45 & \textbf{80.80} & \textbf{88.72} \\
				Ours$_{cross}$ & 94.34 & \textbf{99.75} & \textbf{78.55} & \textbf{86.41} & 79.57 & 88.70 \\ \hline										
			\end{tabular}										
	\end{table}}
\end{center}

		\begin{table}[t]
			\small	
			\centering
			\caption{Comparisons of the in-dataset results on FF++, and the cross-dataset results on Celeb-DF datasets. These methods are trained on FF++.}	
			\vspace{-.05in}	
			\label{tab:cross-dataset1}	
			\begin{tabular}{ccc} \hline	
				\multirow{2}{*}{Method} & FF++ & Celeb-DF \\		
				& AUC & AUC \\ \hline		
				Meso4\cite{Mesonet} \textcolor{gray}{\scriptsize{[WIFS18]}} & 84.70 & 54.80 \\		
				MesoInception4\cite{Mesonet} \textcolor{gray}{\scriptsize{[WIFS18]}} & 83.00 & 53.60 \\		
				RNN\cite{RNN} \textcolor{gray}{\scriptsize{[AVSS18]}} & 90.13 & 63.56 \\		
				FWA\cite{FWA} \textcolor{gray}{\scriptsize{[CVPRW19]}} & 80.10 & 56.90 \\		
				Xception\cite{FF++} \textcolor{gray}{\scriptsize{[ICCV19]}} & 99.70 & 65.30 \\		
				Multi-Task\cite{Multi-task} \textcolor{gray}{\scriptsize{[BTAS19]}} & 76.30 & 54.30 \\		
				Capsule\cite{Capsule} \textcolor{gray}{\scriptsize{[ICASSP19]}} & 96.60 & 57.50 \\		
				EfficientNet-B4\cite{EfficientNet} \textcolor{gray}{\scriptsize{[ICML19]}} & 99.70 & 64.29 \\
				F3-Net\cite{F3-Net} \textcolor{gray}{\scriptsize{[ECCV20]}} & 98.10 & 65.17 \\		
				Multi-Attention\cite{Multi-attention} \textcolor{gray}{\scriptsize{[CVPR21]}} & \textbf{99.80} & 67.44 \\		
				LTW\cite{LTW} \textcolor{gray}{\scriptsize{[AAAI21]}} & 98.50 & 64.10 \\		
				FT-TS\cite{FT-TS} \textcolor{gray}{\scriptsize{[TCSVT21]}} & 92.47 & 65.56 \\
				ADD\cite{Liu2021automated} \textcolor{gray}{\scriptsize{[arXiv21]}} & 91.71 & 66.48 \\		
				FInfer\cite{Hu2022AAAI} \textcolor{gray}{\scriptsize{[AAAI22]}} & 95.67 & 70.60 \\ 
				RECCE\cite{RECCE} \textcolor{gray}{\scriptsize{[CVPR22]}} & 99.32 & 68.71 \\ \hline		
				Ours$_{in}$ & 98.04 & 67.62 \\
				Ours$_{cross}$ & 98.15 & \textbf{72.48} \\ \hline	
			\end{tabular}		
	\end{table}

\begin{table}[t]
	\small	
	\centering
	\caption{Comparisons of the cross-dataset results on WildDeepfake, and DFDC-preview datasets. These methods are trained on FF++. }
	\vspace{-.1in}	
	\label{tab:cross-dataset2}				
	\begin{tabular}{ccc}	\hline			
		\multirow{2}{*}{Method} & WildDeepfake & DFDC-preview \\				
		& AUC & AUC \\	\hline			
		Meso4\cite{Mesonet} \textcolor{gray}{\scriptsize{[WIFS18]}} & 59.74 & 59.30 \\				
		RNN\cite{RNN} \textcolor{gray}{\scriptsize{[AVSS18]}} & 67.03  & 59.37 \\				
		FWA\cite{FWA} \textcolor{gray}{\scriptsize{[CVPRW19]}} & 67.35 & 59.49 \\				
		Xception\cite{FF++} \textcolor{gray}{\scriptsize{[ICCV19]}} & 60.54 & 64.29 \\				
		FT-TS\cite{FT-TS} \textcolor{gray}{\scriptsize{[TCSVT21]}} & 59.82 & 59.09 \\				
		FInfer\cite{Hu2022AAAI} \textcolor{gray}{\scriptsize{[AAAI22]}} & 69.46 & 70.39 \\	
		RECCE\cite{RECCE} \textcolor{gray}{\scriptsize{[CVPR22]}} & 64.31 &  69.06 \\ \hline			
		Ours$_{in}$ & 74.32 & 74.85 \\
		Ours$_{cross}$ & \textbf{74.71} & \textbf{75.79} \\ \hline				
	\end{tabular}				
\end{table}

\begin{figure*}[t]
	\centering
	\includegraphics[width=1\textwidth]{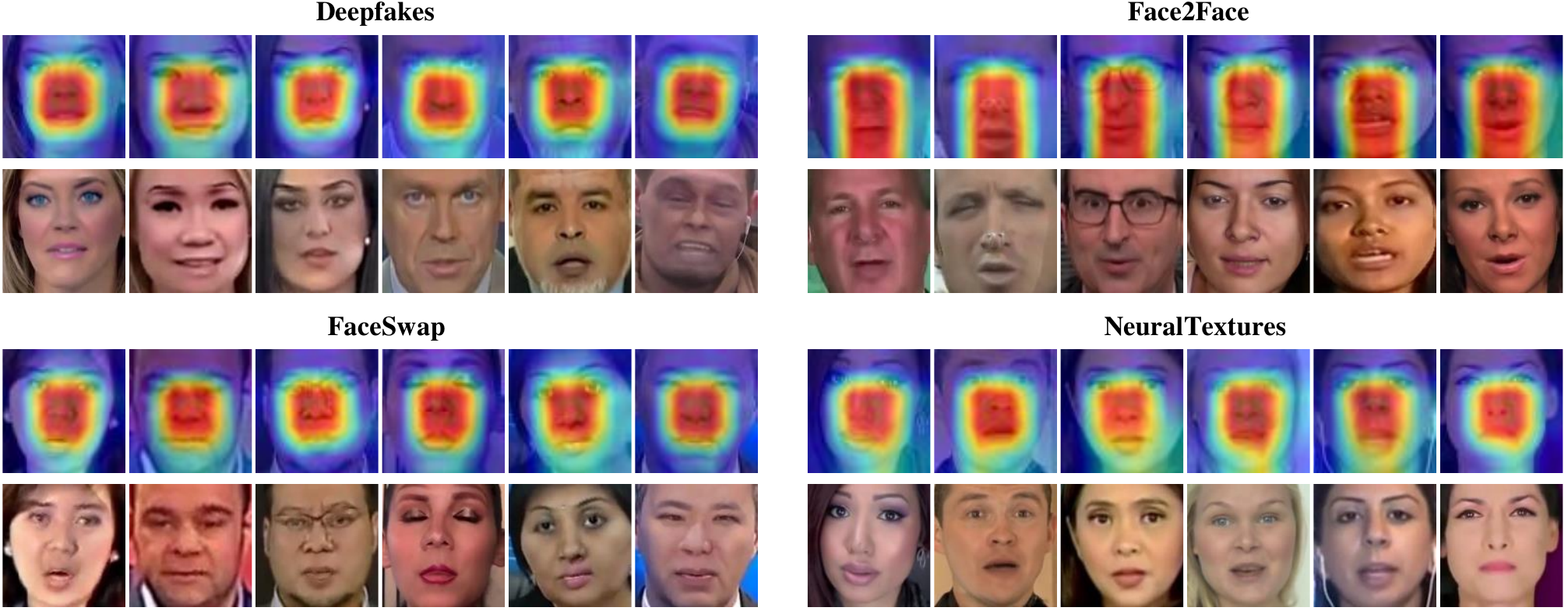}
	\caption{The visualization on FF++. 
	}
	\vspace{-.2in}
	\label{fig:Vis-FF}
\end{figure*}

\subsection{Ablation Study}
In this subsection, we analyze the necessity of some key settings in our method, including search space, search strategy, and cell connection. In the following ablation experiments, we only consider the performance of in-dataset evaluation. The final detection networks are trained on Celeb-DF \cite{Celeb-DF}.

\textbf{Different Search Space.} The forgery-oriented search space plays an important role in our method. 
As shown in Table. \ref{tab:abl-search-space}, our search space is more effective than vanilla CNN in PC-DARTS \cite{Xu2021TPAMI}. Central difference has a positive influence apparently. Various $\theta$ values also perform better than the default value $\theta=0.7$.

\begin{table}[t]
	\small	
	\centering
	\caption{Ablation analysis of different search spaces.}
	\vspace{-.1in}	
	\label{tab:abl-search-space}
	\begin{tabular}{ccc} \hline
		\multirow{2}{*}{Method} & \multicolumn{2}{c}{Celeb-DF} \\
		& Acc & AUC \\ \hline
		vanilla CNN operations (PC-DARTS) \cite{Xu2021TPAMI} & 92.42 & 97.56 \\
		fixed $\theta=0.7$ in CDC & 93.06 & 98.23 \\
		Ours$_{cross}$ & \textbf{94.34} & \textbf{99.75} \\ \hline
	\end{tabular}
\end{table}

\textbf{Different Search Strategy.}
We also compare the effects of the search strategy in Table. \ref{tab:abl-search-strategy}. The proposed method adopts pruning and a novel performance estimation indicator, making the whole algorithm more effective for this forensics task. Concretely, pruning saves nearly half of the search time. The proposed new indicator also contributes for better performance.

\begin{center}
	\setlength{\tabcolsep}{0.8mm}{
\begin{table}[t]
	\small	
	\centering
	\caption{Ablation analysis of different search strategies.}
	\vspace{-.1in}	
	\label{tab:abl-search-strategy}
	\begin{tabular}{cccccc} \hline
		\multirow{2}{*}{Method} & \multirow{2}{*}{pruning} & \multirow{2}{*}{Indicator} & \multirow{1}{*}{Search Time} & \multicolumn{2}{c}{Celeb-DF} \\
		&  &  & (GPU-Days) & Acc & AUC \\ \hline
		w/o pruning+IND & \XSolidBrush & \XSolidBrush & 0.31 & 93.87 & 99.62 \\
		w/o IND & \Checkmark & \XSolidBrush & 0.13 & 94.06 & 99.50 \\
		w/o pruning & \XSolidBrush & \Checkmark & 0.33 & 94.12 & 99.65 \\
		Ours$_{cross}$ & \Checkmark & \Checkmark & 0.13 & \textbf{94.34} & \textbf{99.75} \\ \hline
	\end{tabular}
\vspace{-.1in}
\end{table}}
\end{center}

\textbf{Different Cell Connection.} Since the multiscale information conveys vital clues for forgeries, we gradually downsample the resolutions between cells. As illustrated in Table. \ref{tab:abl-cell}, the number of reduction cells has an impact on performance. As the number of reduction cell increases, the results become better. ``N" represents a normal cell, and ``R" denotes a reduction cell. The proposed C2PN adopts four reduction cells (N$\times$2+R+N$\times$2+R+N$\times$2+R+N$\times$2+R) to progressively downsample feature maps, which obtains the best results.

\begin{table}[t]
	\small	
	\centering
	\caption{Ablation analysis of different cell connections.}
	\vspace{-.1in}	
	\label{tab:abl-cell}		
	\begin{tabular}{ccc} \hline
		\multirow{2}{*}{Connections between cells} & \multicolumn{2}{c}{Celeb-DF} \\
		& Acc & AUC \\ \hline
		N$\times$6+R+N$\times$6+R+N$\times$6 (PC-DARTS) \cite{Xu2021TPAMI} & 93.55 & 97.78 \\
		N$\times$6+R+N$\times$6+R & 93.77 & 98.33 \\
		N$\times$8+R+N$\times$8+R & 94.05  & 98.68 \\
		N$\times$4+R+N$\times$4+R+N$\times$4+R & 94.10 & 95.99  \\
		Ours$_{cross}$ & \textbf{94.34} & \textbf{99.75} \\ \hline
	\end{tabular}
\vspace{-.1in}
\end{table}

\subsection{Visualization Analysis}
\textbf{Detailed Structure of the Best Cells.} The best cells obtained by in-dataset search and cross-dataset search are shown in Fig. \ref{fig:cell} and Fig. \ref{fig:cell-cross-domain}, respectively. The normal cell preserves the resolution, while the reduction cell downsamples the feature maps by half. The operations in both normal cells and reduction cells are obviously different from vanilla CNN, which indicates the significance of central difference convolution. Diverse operations and $\theta$ values are necessary. In Fig. \ref{fig:cell}, the reduction cell adopts dilate convolution to enhance the multiscale information of forged faces. 

\textbf{Activation Maps.}
The Grad-CAM \cite{Grad-CAM} visualizations on FF++ datasets are illustrated in Fig. \ref{fig:Vis-FF}. These activation maps are generated from the outputs of the fourth reduction cell in the final detection networks.
As we can see, most response areas are in the face regions, especially in the nose and the mouth. The clues of different forgery types locate in distinct regions. For example, the activations of Face2Face always cover the whole face, including mouth areas. The activations of FaceSwap are usually concentrated in the central facial regions. 

\section{Conclusion}
In this paper, we propose an end-to-end face forgery detection framework based on NAS, which can automatically design suitable neural networks for this task. First, we create a forgery-oriented search space to find appropriate operations for deepfake detection. Then, we take the actual estimation error as guidance to improve the search generalizability.
Besides ordinary search manners, we also consider cross-dataset search to obtain a more general neural network architecture. 
Eventually, we connect the cells obtained by NAS with skip connections in a cascaded pyramid way. Extensive experimental results demonstrate the proposed method achieves competitive performance compared with state-of-the-art approaches.


{\small
\bibliographystyle{ieee_fullname}
\bibliography{NAS-DFD}
}

\end{document}